\documentclass[runningheads]{llncs}
\usepackage{floatrow}
\usepackage{booktabs}
\usepackage{cite}
\usepackage{graphicx}
\usepackage{multirow}
\usepackage[table,xcdraw]{xcolor}
\usepackage[dvipsnames]{xcolor}
\usepackage{booktabs}
\usepackage{makecell}
\usepackage{threeparttable}
\usepackage{caption}
\usepackage{subcaption}
\usepackage{amsmath}
\usepackage{pifont}

\usepackage{array}
\newcolumntype{L}[1]{>{\raggedright\let\newline\\\arraybackslash\hspace{0pt}}m{#1}}
\newcolumntype{C}[1]{>{\centering\let\newline\\\arraybackslash\hspace{0pt}}m{#1}}
\newcolumntype{R}[1]{>{\raggedleft\let\newline\\\arraybackslash\hspace{0pt}}m{#1}}

\begin{document}
\title{Can Coding Agents Build Robust Baselines? \\ A Skill-Based Approach for Automating the Medical Imaging Model-Development Pipeline}
\titlerunning{Can Coding Agents Build Robust Baselines?}
%

\author{Eugenia~Moris\inst{1}
\and
José~Ignacio~Orlando\inst{1,2}
}
\authorrunning{Moris and Orlando}

\institute{Arionkoder LLC, Boston, MA 02114, United States \email{eugenia.moris@arionkoder.com} \and
Yatiris, PLADEMA, UNICEN-CONICET, Tandil, Buenos Aires, Argentina
}


\maketitle              
\begin{abstract}
Developing competitive deep learning baselines for medical imaging remains a highly iterative process requiring literature review, implementation, experimentation, and expert refinement. Existing automation approaches typically optimize isolated components, such as architecture search or hyperparameter tuning, rather than the complete baseline development process. We present an  agentic AI Scientist workflow that combines literature-guided reasoning, automated code generation, and hypothesis-driven experimentation to generate competitive baseline models for medical imaging challenges. The framework is evaluated on four public benchmarks spanning segmentation, classification, and detection. Across all tasks, the Experimentation Pipeline consistently improves validation performance, achieving competitive leaderboard results, including 6th place on both PUMA tracks (15 teams) and 31st place on MILK10k (125 teams). On MIDOG25, the resulting model also demonstrates strong domain generalization across scanners, tumor types, and species. Using the same workflow across all challenges without task-specific redesign, we demonstrate that skill-based, literature-guided agentic workflows can substantially reduce the engineering effort required to develop competitive medical imaging baselines.

\end{abstract}

\section{Introduction}
Deep learning has become the dominant paradigm for medical image analysis, achieving strong performance in segmentation, classification, and detection \cite{litjens2017survey}. However, developing a competitive baseline for a new problem remains an iterative and expertise-intensive process. It requires numerous interdependent decisions spanning preprocessing, model architecture, optimization, data augmentation, loss design, and evaluation \cite{sculley2015hidden,kreuzberger2023mlops,litjens2017survey,isensee2021nnunet}. These decisions require both clinical and machine learning expertise, making baseline development costly, time-consuming, and difficult to scale \cite{esteva2021guide,kelly2019key,feurer2015autosklearn}.

Previous efforts have addressed isolated stages of this workflow, including hyperparameter optimization \cite{feurer2015autosklearn}, neural architecture search \cite{zoph2017neural,tan2019efficientnet}, and automated data augmentation \cite{cubuk2019autoaugment,cubuk2020randaugment}. More recently, LLM-based agents have extended automation to literature analysis, code generation, experiment execution, and scientific workflows \cite{lwang2023survey,wang2023voyager,lu2024aiscientist,lu2025agentlaboratory,chan2025mlebench}. Nevertheless, existing approaches either optimize isolated stages of model development or provide general-purpose research assistants rather than workflows tailored to medical imaging \cite{feurer2015autosklearn,cubuk2019autoaugment,lu2024aiscientist}. We instead investigate whether literature-guided planning, automated implementation, and evidence-driven experimentation can be integrated into a unified workflow for automatically developing competitive medical imaging baselines with human supervision.

Experienced researchers review the literature before allocating computational resources, using prior evidence to understand the clinical problem, identify promising approaches, and avoid uninformative experiments. Motivated by this process, we propose an agentic AI Scientist workflow that combines literature-guided planning, automated implementation, and hypothesis-driven experimentation through reusable skills. A persistent experiment tree records both positive and negative experimental evidence, improving transparency and reproducibility.
We evaluate the framework on four public medical imaging challenges spanning segmentation, classification, and object detection. Without task-specific redesign, the workflow adapts to heterogeneous tasks, achieves competitive leaderboard performance, and produces strong baselines within 2--7 days. Together, these results demonstrate that literature-guided, evidence-driven AI Scientist workflows can rapidly produce competitive and reproducible medical imaging baselines across diverse tasks.

\section{Methods}



\begin{figure}[t]
    \centering
    \includegraphics[width=\textwidth]{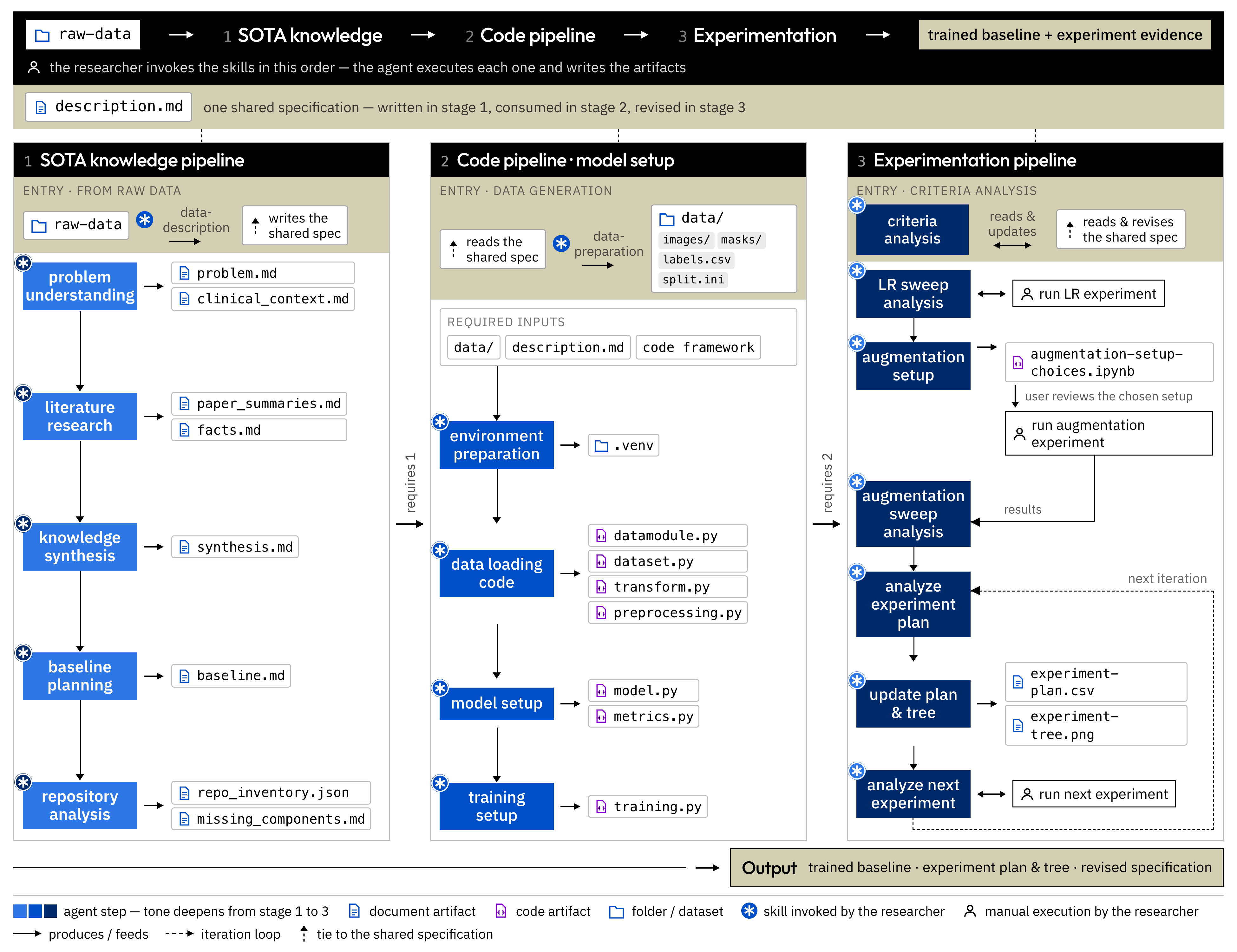}
    \caption{Overview of the proposed skill-based agentic framework. Raw inputs are converted into a structured task description and a standardized dataset. Three phases then synthesize task-specific knowledge, construct an executable modeling pipeline, and iteratively select, execute, and analyze experiments while maintaining a persistent experiment plan and tree. The task description persists across all phases and the researcher invokes each skill in order, reviewing and running the experiments the framework proposes.}
    \label{fig:overview}
\end{figure}

Figure~\ref{fig:overview} presents the proposed AI Scientist workflow. Starting from a labeled dataset, a Data Description skill generates a structured task specification, while a Data Preparation skill converts the original files into a standardized format used by the subsequent pipelines.

The workflow consists of three sequential pipelines. The \emph{State-of-the-Art (SOTA) Knowledge Pipeline} reviews the literature to generate an initial development plan and identify reusable framework components. The \emph{Model Setup Pipeline} automatically implements the required data, model, optimization, and evaluation modules. Finally, the \emph{Experimentation Pipeline} iteratively proposes, executes, and analyzes experiments, updating a persistent experiment plan and experiment tree that guide subsequent decisions within an available computational budget.

Each pipeline is composed of independent LLM-driven skills with explicit inputs and outputs. The resulting artifacts provide a transparent and reproducible record of the complete baseline-development process while keeping the researcher in the loop through lightweight review checkpoints throughout the workflow.



\subsection{SOTA Knowledge Pipeline}

The SOTA Knowledge Pipeline (Figure \ref{fig:overview}, left) transforms a raw dataset into an initial development plan by combining problem characterization, literature analysis, and repository inspection. First, the framework analyzes the task from both clinical and machine learning perspectives, identifying the key challenges, dataset characteristics, evaluation protocol, and potential sources of variability. These observations define the research questions used to guide the subsequent literature review.

Relevant publications and challenge reports are then synthesized to compare evidence across the main design axes, including model architecture, optimization, data augmentation, loss functions, and evaluation. Based on this evidence, the pipeline generates an initial development plan that specifies the baseline configuration. Uncertain or conflicting recommendations are explicitly formulated as experimental hypotheses for the Experimentation Pipeline, while repository analysis identifies the components that can be reused and those requiring implementation.

\subsection{Model Setup Pipeline}
\label{subsec:model-setup}

The Model Setup Pipeline (Figure \ref{fig:overview}, middle) transforms the literature-guided development plan into an executable training framework. Based on the generated plan and repository analysis, it configures the execution environment, implements the dataset-specific loading and preprocessing modules, and instantiates the model architecture, optimization strategy, loss function, and evaluation metrics. The resulting components are integrated into the training framework, producing a reproducible baseline implementation that serves as the starting point for the Experimentation Pipeline. Data augmentation is intentionally excluded, as its policy and parameters are optimized during experimentation.

\subsection{Experimentation Pipeline}
\label{subsec:experimentation}

The Experimentation Pipeline (Figure~\ref{fig:overview}, right) transforms the initial development plan into a competitive baseline through iterative, evidence-driven experimentation. Rather than optimizing a predefined hyperparameter space, it follows a scientific workflow that defines evaluation criteria, tests hypotheses, interprets results, and prioritizes subsequent experiments by expected information gain within a limited computational budget. Evaluation begins by establishing the challenge metric together with secondary metrics for clinically relevant or underrepresented classes. Initial experimental evidence is collected through learning-rate calibration and RandAugment-based policy optimization \cite{cubuk2020randaugment}. Rather than searching unrestricted augmentation parameters, the pipeline first estimates clinically plausible parameter ranges for each candidate transformation using image-based inspection and literature-derived constraints, ensuring that only realistic image variations are explored during policy optimization. Subsequent experiments are generated through a structured process: rule-based logic selects the next component to evaluate, reusable templates define the experiment structure, and the LLM instantiates the hypothesis, configuration, and validation criterion, prioritizing literature-supported alternatives when available. Each experiment modifies a single design decision whenever possible, and completed experiments update a persistent experiment plan and tree that guide future iterations by recording accepted and rejected evidence.

\section{Experimental Setup}

\subsection{Implementation details}
\label{subsec:framework}

The framework is orchestrated through Claude Code \cite{Claude2026}, with all reasoning and code generation performed by Claude Sonnet 4. Each skill is implemented as a task-specific prompt template defining its objective, inputs, outputs, and operational constraints. Rather than relying on a single monolithic prompt, the workflow decomposes the development process into specialized skills that exchange structured Markdown, YAML, and CSV artifacts.
The implementation is based on a shared PyTorch Lightning \cite{falcon2019pytorch} framework providing the common training infrastructure. The Model Setup Pipeline automatically generates the task-specific components—including the data module, model, loss functions, augmentations, optimizer, and evaluation configuration—within this reusable framework.
Experiments were executed on two hardware platforms. PUMA tissue segmentation was developed on Apple Silicon MPS backend. PUMA nuclei detection and MILK10k were trained on an NVIDIA RTX 4090 (24 GB), while MIDOG25 initially used MPS before transitioning to the RTX 4090 for foundation-model experiments requiring additional GPU memory.

Although the workflow automates literature synthesis, code generation, experiment planning, and configuration, human supervision remains part of the development process. After each stage, the researcher reviews the generated artifacts, validates the proposed implementation, launches the suggested experiments, and monitors their execution. Manual modifications to the generated modeling code were not required in the reported experiments. Instead, the reusable framework enables the agent to generate only the task-specific components while reusing a common training, optimization, logging, and evaluation infrastructure. The researcher's role shifts from implementing and configuring models to supervising the development process and validating the generated evidence, preserving human oversight while substantially reducing engineering effort.

\subsection{Materials and evaluation}
\label{subsec:eval}

\begin{table}[t]
\centering
\caption{Summary of the benchmark challenges used to evaluate the workflow.}
\label{tab:datasets}
\small
\resizebox{\linewidth}{!}{%
\begin{tabular}{lcccc}
\toprule
\textbf{Challenge} & \textbf{Task} & \textbf{Train/Val/Test} & \textbf{Metric} & \textbf{Evaluation} \\
\midrule
PUMA &
\makecell{Segmentation + Detection} &
174 / 31 / Hidden &
Dice / Macro-F1 &
Leaderboard \\
MILK10k &
Classification &
4454 / 786 / Hidden &
Macro-F1 &
Leaderboard \\
MIDOG25 &
Detection &
8626 / 1747 / 1564 &
Object F1 &
Held-out test \\
\bottomrule
\end{tabular}}
\end{table}

The proposed workflow was evaluated on four benchmark tasks derived from three public medical imaging challenges covering semantic segmentation, multiclass classification, and object detection (Table~\ref{tab:datasets}). The workflow first searches for patient-, specimen-, or slide-level identifiers and generates group-aware data splits whenever available. Otherwise, it uses reproducible image-level random splitting. As PUMA and MILK10k provide no such metadata, we created 85/15 image-level splits (seed 42). MIDOG25 instead provides whole-slide identifiers, enabling grouped train, validation, and held-out test splits.

For all challenges, model selection relied exclusively on validation performance using the official challenge metrics. For MIDOG25, we report both Macro-$F_1$ and $F_{1,\mathrm{AMF}}$, the F1-score of the atypical mitotic figure class. The Experimentation Pipeline recorded the validation metric after learning-rate calibration, augmentation optimization, and experiment exploration. Across all three stages, the total number of training runs was capped at 100 per challenge. Final models for PUMA and MILK10k were evaluated through the official challenge servers, whereas MIDOG25 was evaluated on the independent held-out test split. We additionally compare our results with the most similar published approach and, when available, the top-ranked challenge submission.

\section{Results and Discussion}


\begin{table}[t]
\centering
\caption{Performance across challenges with open leaderboard evaluations. Validation scores are reported after learning-rate (LR) selection, data augmentation, and experiment exploration. Test scores include results of our final baseline vs. the most similar method submitted and the leading model.}
\label{tab:challenges_results}

\begin{threeparttable}
\scriptsize
\renewcommand{\arraystretch}{1.15}

\resizebox{\linewidth}{!}{%
\begin{tabular}{@{}llcccccccc@{}}
\toprule
\multirow{2}{*}{\textbf{Challenge}} &
\multirow{2}{*}{\textbf{Metric}} &
\multicolumn{3}{c}{\textbf{Validation}} &
\multicolumn{4}{c}{\textbf{Test}} &
\multirow{2}{*}{\makecell{\textbf{Time to model}\\\textbf{(days)}\tnote{\textdagger}}}
\\
\cmidrule(lr){3-5}
\cmidrule(lr){6-9}
&
&
\makecell{\textbf{LR}\\\textbf{selection}} &
\makecell{\textbf{Data}\\\textbf{augmentation}} &
\makecell{\textbf{Experiment}\\\textbf{exploration}} &
\makecell{\textbf{Final baseline}\\\textbf{(ours)}} &
\makecell{\textbf{Similar}\\\textbf{method}} &

\makecell{\textbf{First-place}\\\textbf{model}} &
\textbf{Rank} &
\\
\midrule

\makecell[l]{PUMA\\Tissue segmentation}
&
Macro Dice
&
0.672
&
0.675
&
0.732
&
\textbf{0.691}
&
\makecell{0.554\\\textcolor{green!45!black}{\scriptsize $(+0.137)$}}

&
\makecell{0.783\\\textcolor{red!70!black}{\scriptsize $(-0.092)$}}
&
6/15
&
2
\\

\addlinespace[2pt]

\makecell[l]{PUMA\\Nuclei detection}
&
Macro-F1
&
0.459
&
0.467
&
0.717
&
\textbf{0.624}
&
\makecell{0.586\\\textcolor{green!45!black}{\scriptsize $(+0.038)$}}

&
\makecell{0.658\\\textcolor{red!70!black}{\scriptsize $(-0.034)$}}
&
6/15
&
4
\\

\addlinespace[2pt]

\makecell[l]{MILK10k\\ISIC Challenge}
&
Macro-F1
&
0.474
&
0.493
&
0.528
&
\textbf{0.490}
&
\makecell{0.515\tnote{*}\\\textcolor{red!70!black}{\scriptsize $(-0.025)$}}

&
\makecell{0.609\tnote{*}\\\textcolor{red!70!black}{\scriptsize $(-0.119)$}}
&
31/125\tnote{*}
&
5
\\

\addlinespace[2pt]

\makecell[l]{MIDOG25\\Atypical mitosis classification}
&
F1
&
0.757
&
0.791
&
0.855
&
--- $\ddagger$
&
0.882

&
0.907
&
---
&
7
\\

\bottomrule
\end{tabular}%
}

\begin{tablenotes}[flushleft]
\scriptsize
\item[*] Among methods that do not use external datasets.
\item[\textdagger] Elapsed project duration measured from the first executed training run to the final selected \\ baseline; not researcher working hours.
\item[$\ddagger$] A direct comparison is not possible because our model was evaluated on a different test set.
\end{tablenotes}
\end{threeparttable}
\end{table}

Before experimentation, the SOTA Knowledge Pipeline generated an initial development plan by synthesizing evidence from the literature and challenge documentation. The resulting plan specified the initial model architecture, optimization strategy, input resolution, loss function, augmentation policy, evaluation protocol, and prioritized experimental hypotheses.

Across all challenges, the Experimentation Pipeline validated many literature-derived decisions—including optimization strategies, augmentation policies, and several training configurations—while revising others, most notably the model architecture.
These findings suggest that literature-guided planning provides a strong starting point, while systematic experimentation remains essential for adapting the solution to the characteristics of a specific dataset.

Table~\ref{tab:challenges_results} quantifies the effect of the refinement process. Validation performance improved after each stage of the Experimentation Pipeline across all four tasks, with the largest gains obtained through experiment exploration. The final models also performed competitively on independent leaderboard evaluations, ranking 6th of 15 teams in both PUMA tracks and 31st of 125 submissions on MILK10k. Although leaderboard comparisons are not controlled, these results suggest that the proposed workflow generalizes across diverse medical imaging tasks.



\begin{figure}[t]
  \centering
  \includegraphics[width=\textwidth]{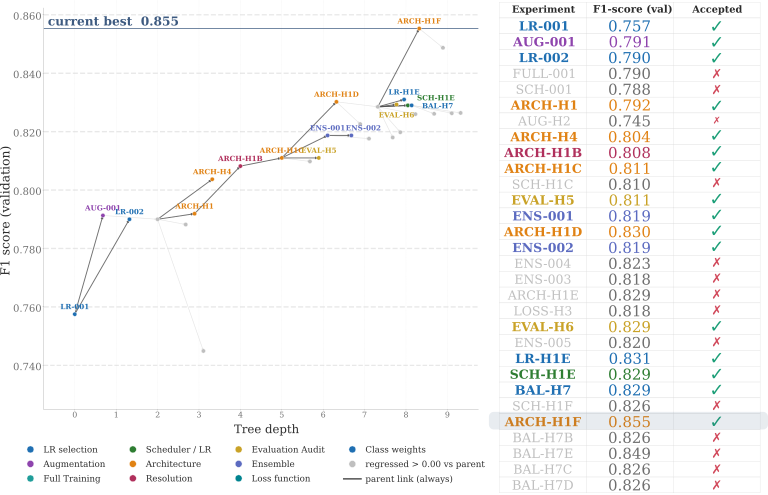}
  \caption{Experimentation workflow on MIDOG25. Left: experiment tree showing validation $F_1$ progression, with nodes colored by decision category and gray nodes indicating rejected experiments. Right: chronological sequence of executed experiments. The final accepted configuration achieved the highest validation score ($F_1=0.855$).}
  \label{fig:midog25_progression}
\end{figure}

Figure~\ref{fig:midog25_progression} illustrates how the Experimentation Pipeline incrementally refines the initial development plan through evidence-driven experimentation. Across 30 MIDOG25 experiments, the best validation $F_1$ improved from 0.757 to 0.855. The framework evaluated hypotheses spanning optimization, data augmentation, architecture, input resolution, and training strategies while recording both successful and unsuccessful experiments. Rather than being discarded, rejected experiments are preserved as negative evidence that guides subsequent decisions. Overall, 17 experiments were accepted and 13 were rejected, allowing the search to branch from the strongest validated configuration while avoiding previously explored directions. Unlike conventional experiment tracking systems such as MLflow and Weights \& Biases, which primarily organize and record experimental artifacts and metadata, the proposed experiment tree preserves structured positive and negative evidence to directly inform future experimentation \cite{zaharia2018mlflow,biewald2020wandb}.

         

\begin{table}[t]
\centering
\caption{Evaluation of MIDOG25. Left: overall validation and test performance. Right: summary of domain generalization across heterogeneous domains. }
\label{tab:midog25_generalization}

\begin{minipage}[t]{0.3\textwidth}
\centering
\textbf{Overall}\\[2pt]
\small
\begin{tabular}{lcc}
\toprule
Metric & Val. & Test \\
\midrule
Macro-$F_1$ & 0.8553 & 0.8473 \\
$F1_{AMF}$ & 0.7579 & 0.7457 \\
\bottomrule
\end{tabular}
\end{minipage}
\hfill
\begin{minipage}[t]{0.6\textwidth}
\centering
\textbf{Domain Generalization}\\[2pt]
\small
\begin{tabular}{lccc}
\toprule
Domain & Min & Max & Mean \\
\midrule
Species & 0.731 & 0.764 & 0.748 \\
Tumor types & 0.633 & 0.880 & 0.771 \\
Scanners$^\dagger$ & 0.694 & 0.807 & 0.756 \\
\bottomrule
\end{tabular}

\vspace{2pt}
\raggedright
\footnotesize
$^\dagger$ Excluding Hamamatsu S360 scanner (1 AMF).
\end{minipage}

\end{table}

Table~\ref{tab:midog25_generalization} summarizes the domain generalization of the automatically generated MIDOG25 baseline. Validation and held-out test performance remained consistent, with similar results across species, scanners, and tumor types despite substantial biological and acquisition variability. Although evaluated on a held-out split derived from the released challenge data, these findings suggest that literature-guided planning and evidence-driven experimentation can produce robust baselines that generalize across heterogeneous medical imaging domains.

\subsection{Limitations}

This study evaluates the integrated workflow rather than the contribution of its individual components. We therefore do not isolate the effect of the SOTA Knowledge Pipeline, the Model Setup Pipeline, or the Experimentation Pipeline through ablation studies. In addition, experiments were performed using a single random seed, no confidence intervals are reported, and we did not compare against human-developed baselines, conventional hyperparameter optimization, or existing AutoML systems under identical computational budgets. Consequently, the reported results should be interpreted as evidence for the effectiveness of the complete workflow rather than of its individual modules. Future work should benchmark against established AutoML and augmentation-search methods (e.g., auto-sklearn, auto-augment).

Similar patterns emerged across the evaluated challenges. Architecture exploration consistently produced the largest performance gains, whereas loss-based class reweighting and learning-rate schedulers were never retained. Performance on underrepresented classes was instead improved through oversampling in PUMA and class-specific decision thresholds in MILK10k. Although these are descriptive observations rather than controlled ablations, they show that the experiment tree captures recurring optimization patterns across tasks.

Finally, the workflow operates within a predefined orchestration strategy and reusable training framework. Although the agent iteratively refines development decisions using experimental evidence, it cannot autonomously redesign the workflow, revisit the literature in response to unexpected findings, or introduce new development stages. Enabling more adaptive scientific reasoning while preserving transparency, reproducibility, and human oversight remains an important direction for future AI Scientist systems. Future studies will investigate more autonomous scientific reasoning inspired by recent AI Scientist systems \cite{lu2024aiscientist,lu2025agentlaboratory}.

\section{Conclusion}

We presented an agentic framework that helps develop competitive baseline models for medical imaging challenges. The workflow combines literature-guided planning, automated implementation, and evidence-driven experimentation to transform raw data into a trained baseline with lightweight human oversight.

We evaluated the framework on four public medical imaging challenges spanning segmentation, classification, and detection. Across all tasks, the Experimentation Pipeline consistently improved validation performance, while the resulting automatically generated baselines achieved competitive leaderboard performance. The MIDOG25 results further demonstrated that the generated baseline generalized without requiring dedicated domain-adaptation strategies.

These findings demonstrate that a structured AI Scientist workflow can automate a substantial portion of the machine learning development process for medical imaging while maintaining competitive performance across diverse tasks. More broadly, we believe this study is a step toward AI systems that support researchers throughout the model development lifecycle — letting experts focus on scientific reasoning, hypothesis generation, and clinical interpretation rather than repetitive implementation — and that skill-based AI Scientists represent a practical path toward scalable, reproducible machine learning research in medical imaging.



\begin{credits}

\subsubsection{\ackname}
The authors thank Arionkoder for supporting this research and providing the computational resources used in this study.

\subsubsection{\discintname}
The authors declare that they have no competing interests relevant to the content of this article.

\end{credits}

\bibliographystyle{plain}
\bibliography{bibliography}

@inproceedings{sculley2015hidden,
  title     = {Hidden Technical Debt in Machine Learning Systems},
  author    = {Sculley, D. and others},
  booktitle = {{NeurIPS}},
  volume    = {28},
  year      = {2015}
}

@article{kreuzberger2023mlops,
  title   = {Machine Learning Operations ({MLOps}): Overview, Definition, and Architecture},
  author  = {Kreuzberger, Dominik and others},
  journal = {IEEE Access},
  volume  = {11},
  pages   = {31866--31879},
  year    = {2023}
}

@article{litjens2017survey,
  title   = {A Survey on Deep Learning in Medical Image Analysis},
  author  = {Litjens, Geert and others},
  journal = {Medical Image Analysis},
  volume  = {42},
  pages   = {60--88},
  year    = {2017}
}

@article{isensee2021nnunet,
  title   = {{nnU-Net}: A Self-Configuring Method for Deep Learning-Based Biomedical Image Segmentation},
  author  = {Isensee, Fabian and others},
  journal = {Nature Methods},
  volume  = {18},
  number  = {2},
  pages   = {203--211},
  year    = {2021}
}

@article{esteva2021guide,
  title   = {Deep Learning-Enabled Medical Computer Vision},
  author  = {Esteva, Andre and others},
  journal = {npj Digital Medicine},
  volume  = {4},
  number  = {1},
  pages   = {5},
  year    = {2021}
}

@article{kelly2019key,
  title   = {Key Challenges for Delivering Clinical Impact with Artificial Intelligence},
  author  = {Kelly, Christopher J. and others},
  journal = {BMC Medicine},
  volume  = {17},
  number  = {1},
  pages   = {195},
  year    = {2019}
}

@inproceedings{feurer2015autosklearn,
  title     = {Efficient and Robust Automated Machine Learning},
  author    = {Feurer, Matthias and others},
  booktitle = {{NeurIPS}},
  volume    = {28},
  year      = {2015}
}

@inproceedings{cubuk2019autoaugment,
  title     = {AutoAugment: Learning Augmentation Strategies from Data},
  author    = {Cubuk, Ekin D. and others},
  booktitle = {{CVPR}},
  pages     = {113--123},
  year      = {2019}
}

@inproceedings{cubuk2020randaugment,
  title     = {RandAugment: Practical Automated Data Augmentation with a Reduced Search Space},
  author    = {Cubuk, Ekin D. and others},
  booktitle = {{CVPRW}},
  pages     = {702--703},
  year      = {2020}
}

@article{lu2025agentlaboratory,
  title   = {Agent Laboratory: Using {LLM} Agents as Research Assistants},
  author  = {Schmidgall, Samuel and others},
  journal = {arXiv:2501.04227},
  year    = {2025}
}

@article{wang2023voyager,
  author = {Wang, Guanzhi and others},
  title = {Voyager: An Open-Ended Embodied Agent with Large Language Models},
  journal = {arXiv:2305.16291},
  year = {2023},
  url = {https://arxiv.org/abs/2305.16291}
}

@inproceedings{zoph2017neural,
  title={Neural Architecture Search with Reinforcement Learning},
  author={Zoph, Barret and Le, Quoc},
  booktitle={ICLR},
  year={2017}
}

@inproceedings{tan2019efficientnet,
  title={{EfficientNet}: Rethinking Model Scaling for Convolutional Neural Networks},
  author={Tan, Mingxing and Le, Quoc},
  booktitle={ICML},
  year={2019}
}

@article{lwang2023survey,
  title={A Survey on Large Language Model based Autonomous Agents},
  author={Wang, Lei and others},
  journal={arXiv preprint arXiv:2308.11432},
  year={2023}
}

@article{lu2024aiscientist,
  title={The {AI} Scientist: Towards Fully Automated Open-Ended Scientific Discovery},
  author={Lu, Chris and others},
  journal={arXiv preprint arXiv:2408.06292},
  year={2024}
}

@article{zaharia2018mlflow,
  title={Accelerating the Machine Learning Lifecycle with {MLflow}},
  author={Zaharia, Matei and others},
  journal={IEEE Data Engineering Bulletin},
  year={2018}
}

@misc{biewald2020wandb,
  author = {Lukas Biewald},
  title = {Experiment Tracking with Weights and Biases},
  year = {2020},
  howpublished = {\url{https://www.wandb.com/}}
}

@inproceedings{chan2025mlebench,
  title={{MLE-Bench}: Evaluating Machine Learning Agents on Machine Learning Engineering},
  author={Chan, Kyle and others},
  booktitle={International Conference on Learning Representations (ICLR)},
  year={2025}
}

@software{Claude2026,
  author = {{Anthropic}},
  title = {Claude},
  version = {3.7 Sonnet},
  year = {2026},
  url = {https://claude.ai},
  note = {Accessed: 2026-07-30}
}

@article{falcon2019pytorch,
  title={PyTorch Lightning},
  author={Falcon, WA},
  journal={GitHub},
  volume={3},
  year={2019},
  note={https://github.com/Lightning-AI/pytorch-lightning}
}


\end{document}